\documentclass[conference]{IEEEtran}
\IEEEoverridecommandlockouts
\usepackage{cite}
\usepackage{amsmath,amssymb,amsfonts}
\usepackage{algorithmic}
\usepackage{multirow}
\usepackage{subcaption}
\usepackage{graphicx}
\usepackage{stmaryrd}

\usepackage{stfloats}

\usepackage{textcomp}
\usepackage{xcolor}
\def\BibTeX{{\rm B\kern-.05em{\sc i\kern-.025em b}\kern-.08em
    T\kern-.1667em\lower.7ex\hbox{E}\kern-.125emX}}
\begin{document}

\title{Residual or Gate? Towards Deeper Graph Neural Networks for Inductive Graph Representation Learning}

\author{\IEEEauthorblockN{1\textsuperscript{st} Binxuan Huang}
\IEEEauthorblockA{\textit{School of Computer Science} \\
\textit{Carnegie Mellon University }\\
Pittsburgh, United States \\
binxuanh@cs.cmu.edu}
\and
\IEEEauthorblockN{2\textsuperscript{nd} Kathleen M. Carley}
\IEEEauthorblockA{\textit{School of Computer Science} \\
\textit{Carnegie Mellon University}\\
Pittsburgh, United States \\
kathleen.carley@cs.cmu.edu}
}

\maketitle

\begin{abstract}
In this paper, we study the problem of node representation learning with graph neural networks. We present a graph neural network class named recurrent graph neural network (RGNN), that address the shortcomings of prior methods. By using recurrent units to capture the long-term dependency across layers, our methods can successfully identify important information during recursive neighborhood expansion. In our experiments, we show that our model class achieves state-of-the-art results on three benchmarks: the Pubmed, Reddit, and PPI network datasets. Our in-depth analyses also demonstrate that incorporating recurrent units is a simple yet effective method to prevent noisy information in graphs, which enables a deeper graph neural network.
\end{abstract}


\section{Introduction}
Graphs are universal models of objects and their pairwise relationships. We can view many data in the form of graphs, including social networks, protein interactions, paper citations. But unlike sequence data or grid data, it is hard to express and exploit graph information in many machine learning tasks. Recently substantial efforts have been made to learn expressive structure information in graphs \cite{perozzi2014deepwalk, grover2016node2vec, tang2015line}. 

Generally, there are two ways to represent graph information in recent literature. One is network embedding, where a node's neighbourhood information is condensed into a vector representation. Then we can use these node vector representations for downstream tasks like node classification \cite{bhagat2011node} and link prediction \cite{lu2011link}. A typical example is DeepWalk \cite{perozzi2014deepwalk}. However, these types of embedding methods assume the whole graph is given in advance. If new nodes are added afterwards, then we have to re-train embedding vectors for the whole new graph. Because the re-trained embedding space may not align with the original one, we have to re-train all downstream tasks, which is not practical in the real world. Another way is using graph neural networks (GNN), which have shown great success in learning node representations \cite{kipf2016semi, hamilton2017inductive}. Graph neural networks are deep learning-based methods that operate on graphs. At each layer, GNNs aggregate information from neighbourhoods and generate hidden states for each node. Because GNNs do not require a fixed graph, we can easily apply them to new graphs on the fly, which is suitable for the inductive setting.

In recent proposed GNNs, there is a common drawback that training becomes extremely difficult when models become deeper \cite{kipf2016semi}. This is partially due to more layers would also propagate noisy information from expanded neighborhood \cite{kipf2016semi}. Though researchers try to use residual connection \cite{he2016deep} to overcome this issue \cite{kipf2016semi, velivckovic2017graph}, the performance still gets worse with deeper models. In this paper, we will show that using residual connection is not the best option for a deep graph neural network. Rather, incorporating recurrent units in graph neural networks can effectively capture the neighbourhood information while keeping local features unvarnished. In this work, we present a deep graph neural network class named recurrent graph neural networks (RGNN). It uses recurrent units to compress previous neighborhood information into hidden states and successfully captures useful information during recursive neighborhood expansion. 

In our experiments, we systematically evaluate the performance of RGNNs under supervised learning setting as well as unsupervised setting. In our comparative evaluation, we show RGNNs can consistently improve the performance of base GNN models. Our model class achieves state-of-the-art results on three commonly used benchmark datasets. We further compare this neural network class with GNNs with residual connections. Experiments show that RGNNs have better learning capability with the same number of layers and can effectively avoid noisy neighbourhood information. 

The rest of this paper is organized as follows. In Section II, we provide some background knowledge about recurrent units and graph neural networks. The detail of our framework is given in Section III. We present our experiments and model analyses in Section IV. We summarize related work and conclude this work in Section V and VI respectively.

\section{Preliminary}

\subsection{Recurrent Neural Networks}
Recurrent neural networks (RNN) are a class of neural network designed to handle sequence input. At each time step $t$, an RNN takes an input $x_t$ and a previous hidden state $h_{t-1}$, generates a hidden state vector as $h_t=RNN(x_t,h_{t-1})$. The update rule for a vanilla RNN is as follows: 
\begin{align}
h_t = tanh(Wh^{t-1}+Ux_t+b)
\end{align}
where $W, U$ are parameter matrices and $b$ is a bias vector.

People use RNN to capture long term dependency in sequence and compress previous history into the hidden state. However, because a vanilla RNN often faces gradient vanishing and exploding problem, researchers use long short-term memory (LSTM) \cite{hochreiter1997long} or gated recurrent unit (GRU) \cite{cho2014learning} as better alternatives.

For an LSTM unit, given an input vector $x_t$, a previous LSTM cell state $c_{t-1}$ and a previous LSTM hidden state $h_{t-1}$, it updates the current cell state $c_t$ and the hidden state $h_t$ at time step $t$ by the following rules:
\begin{align}
i_t &= \sigma(W_i  x_t+U_i  h_{t-1}+b_i) \\
f_t &= \sigma(W_f x_t+ U_f  h_{t-1}+b_f) \\
o_t &= \sigma(W_o  x_t+U_o  h_{t-1}+b_o)\\
\hat c_t &=  tanh(W_c  x_t+U_c  h_{t-1}+b_c)\\
c_t &= f_t \circ c_{t-1}+i_t\circ \hat c_t\\
h_t &= o_t \circ tanh(c_t)
\end{align}
where $\sigma(\cdot)$ and $tanh(\cdot)$ are the sigmoid function and hyperbolic tangent function respectively. $W_i, U_i, W_f$, $U_f, W_o,U_o, W_c,U_c$ are parameter matrices and $b_i,b_f,b_o,b_c$ are bias vectors to be learned during training. Symbol $\circ$ represents element-wise multiplication. 
$i_t$, $f_t$ and $o_t$ are input gate, forget gate and output gate, which control the information flow. 
$f_t$ is a forget gate which controls how much previous cell state should be kept in current step. The input gate $i_t$ determines how much information should be added from $\hat c_t$ to current cell state. Output gate $o_t$ controls how many information should be exposed to the hidden state. Because LSTM's ability to model longer dependency, it has been widely adopted in many sequence modeling tasks \cite{sundermeyer2012lstm, gers2002learning}.

GRU is another popular variant of RNN. It is similar with LSTM, but with fewer parameters. At step $t$, it updates the hidden state $h_t$ as 
\begin{align}
    z_t &= \sigma(W_zx_t+U_zh_{t-1}+b_z)\\
    r_t &= \sigma(W_rx_t+U_rh_{t-1}+b_r)\\
    \hat h_t &= tanh(W_hx_t+U_h(r_t \circ h_{t-1})+b_h)\\
    h_t &= (1-z_t)\circ h_{t-1}+z_t\circ \hat h_t
\end{align}
where $W_z, U_z, W_r, U_r, W_h,U_h$ are parameter matrices, $b_z, b_r,b_h$ are bias vectors. $r_t$ is a reset gate that decides how much of the past information to forget in $\hat h_t$. $z_t$ is an update gate that determines how much of past information to be passed to the next step. Empirically, GRU performs similarly with LSTM in many sequence modeling tasks \cite{chung2014empirical}.

\subsection{Graph Neural Networks}
The concept of graph neural networks was first introduced in \cite{scarselli2009graph}. Given a graph with adjacent matrix $A\in R^{N\times N}$, the representation $h_i$ for a node $i$ is updated as follows:
\begin{align}
    h_i &= f(x_i,x_{e[i]},h_{n[i]},x_{n_i}) \\
    o_i &= g(h_i,x_i)
\end{align}
where $x_i$, $x_{e[i]}$, $h_{n[i]}$, $x_{n_i}$ are features of node $i$, features of its edges, the states, and the features of its neighbourhood. Function $f$ is a contraction map and are shared across layers. The final representation $h_i$ for node $i$ is a fixed point of $f$. Combining $h_i$ and $x_i$, it outputs label $o_i$ for node $i$. In general, this process can be viewed as features propagation from neighbourhood. 

There are several GNN variants exist in the literature. Kipf and Welling introduce a simplified spectral approach called graph convolutional neural networks (GCN) \cite{kipf2016semi}. They use one-step neighbourhood to update the state of a central node as: 
\begin{align}
    H^{l+1} = \hat D ^{-\frac{1}{2}} \hat A \hat D ^{-\frac{1}{2}} H^l \Theta^l 
\end{align}
where $\hat A=A+I$, $\hat D_{ii}=\sum_j\hat A_{ij}$, $H^l \in R^{N\times C_l}$ is the stacked states for all nodes at layer $l$, $H^0$ is stacked node features $X$, $\Theta^l \in R^{C_l\times C_{l+1}}$ is a filter parameter. $C_l$ is the dimension of hidden states at layer $l$.

Another popular variant is the graph attention network (GAT) \cite{velivckovic2017graph}. Again, a node's state is updated by aggregating its neighbourhood's states. GAT adopted one widely used multi-head attention method in natural language processing to learn important nodes in neighbourhood \cite{vaswani2017attention}. Using K attention heads, GAT update states by 
\begin{align}
    h_i^{l+1} &=\bigparallel_{k=1}^K \sigma(\sum_{j\in n[i]} \alpha^{lk}_{ij}W^{lk}h_j^{l})\\
    \alpha^{lk}_{ij} &= \frac{exp(LeakyReLU({a_k^l}^T[W^{lk}h_i^l||W^{lk}h_j^l]))}{\sum_{u\in n[i]}exp(LeakyReLU({a_k^l}^T[W^{lk}h_i^l||W^{lk}h_u^l]))}
\end{align}
where $\bigparallel$ represents vector concatenation, $\alpha^{lk}_{ij}$ is the attention coefficient of node $i$ to its neighbour $j$ in attention head $k$ at layer $l$. $W^{lk} \in R^{\frac{C_{l+1}}{K}\times C_l}$ is a linear transformation for input states. $\sigma$ denotes a sigmoid function. $a_k^l \in R^{\frac{2C_{l+1}}{K}}$ is an attention context vector learned during training. 

In practise, researchers have observed that deeper GNN models could not improve performance and even perform worse, which is partially due to more layers would also propagate noisy information from expanded neighborhood \cite{kipf2016semi}. A common option is using a residual connection as shown in Eq. \ref{res}, which adds states from lower layer directly to higher layer and avoids the local features getting vanished in higher layers.
\begin{align}
    H^{l+1} = GNN(H^l,A;\Theta^l) + H^{l}
    \label{res}
\end{align}
where $\Theta_l$ is parameter of GNN at layer $l$.

\section{Recurrent Graph Neural Network}

In a GNN model, each layer $l$ can potentially capture information from neighbours with $l$-hops distance. Such deep GNNs could propagate noisy information from the expanded neighbourhood. An intuitive thought would be can we use recurrent units to model long-term dependency across layers. If we take hidden states across layers as a sequence of observations, a recurrent unit with good sequence modeling capability can ideally compress previous graph history into node states and control how much information should be added to new hidden states.

\begin{figure}[!h]
  \centering
  \includegraphics[width=1\linewidth]{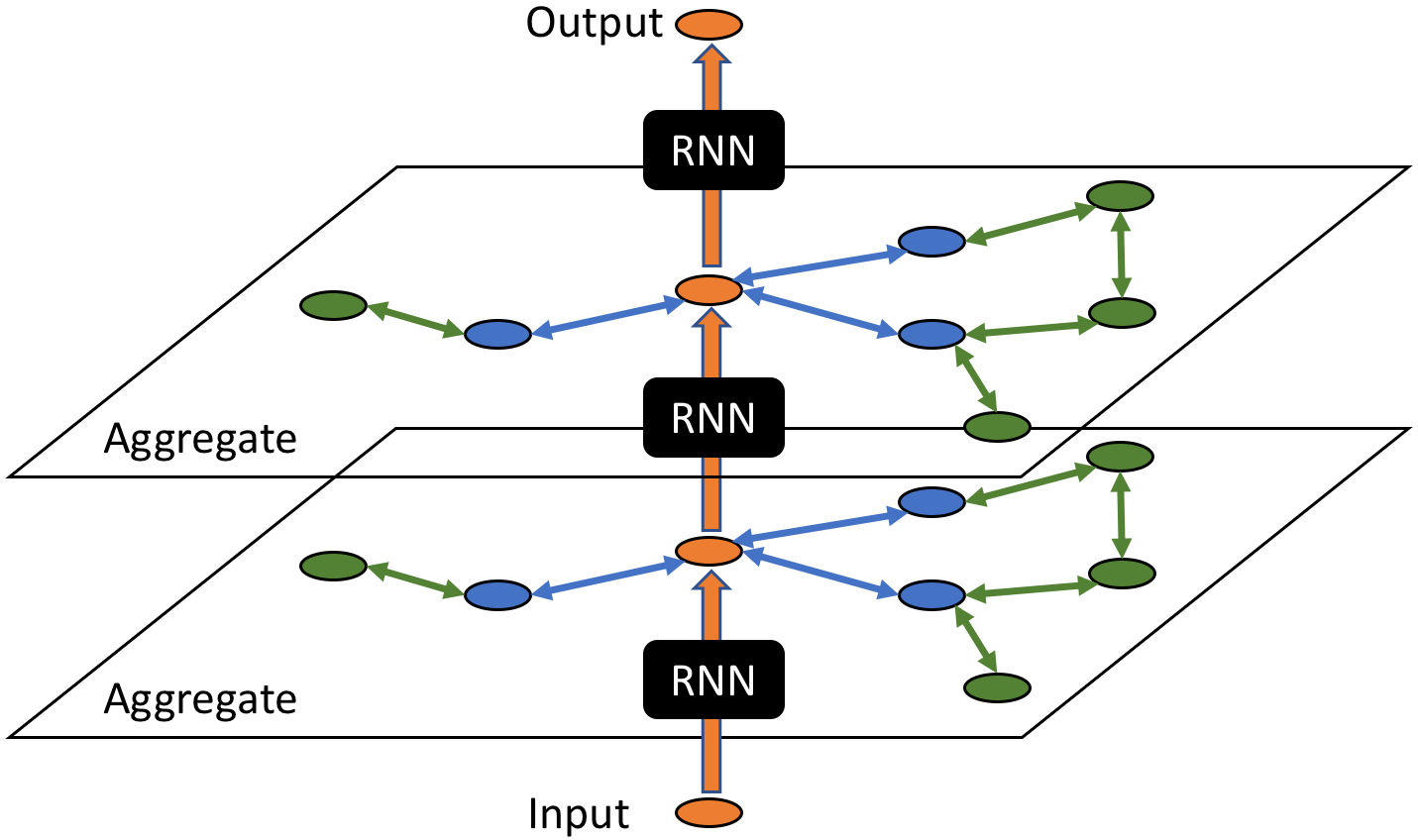}
  \caption{A visual illustration of the feedforward process by a central node in a two-layer RGNN model.}
\label{arch}
\end{figure}

With this intuition, we present the general recurrent graph neural network framework as follows:
\begin{align}
    H^{l+1} &= RNN(GNN(H^l,A;\Theta^l),H^l), \ \ \ \ \ l\ge 0\\
    H^0 &= RNN(W_i X+b_i,0)
\end{align}
where at each layer $GNN(H^l,A; \Theta^l)$ generates new input for an RNN unit, and this RNN unit decides how much information should be added into the next layer. The initial hidden state $H^0$ is generated by feeding node local features into the RNN unit. $W_i,b_i$ are a projection matrix and a bias vector that maps input features into the dimension of hidden states. The feedforward process of a two-layer RGNN model is illustrated in Figure \ref{arch}.

An intuitive view of this RGNN model is that at layer $0$ the hidden state $h^0$ is only dependent on the node's local features, and at each layer $l$ information from $l$-hop neighbourhood is compressed into the hidden state by a recurrent unit. 

Take a graph convolutional neural network with long short-term memory (RGCN-LSTM) for example, it updates node representations at each layer as follows:
\begin{align}
       \hat  X^{l+1} &= \hat D ^{-\frac{1}{2}} \hat A \hat D ^{-\frac{1}{2}} H^l \Theta^l \\
       I^{l+1} &= \sigma(  X^{l+1}W_i+  H^lU_i+[b_i]_N) \\
        F^{l+1} &= \sigma( X^{l+1}W_f+   H^lU_f+[b_f]_N) \\
        O^{l+1} &= \sigma(  X^{l+1}W_o+  H^lU_o+[b_o]_N)\\
        \hat C^{l+1} &=  tanh( X^{l+1}W_c+  H^lU_c+[b_c]_N)\\
        C^{l+1} &= F^{l+1} \circ C^l+I^{l+1}\circ \hat C^{l+1}\\
        H^{l+1} &= O^{l+1} \circ tanh(C^{l+1})
\end{align}

where $[b]_N$ represents stacking bias vector $b\in R^C$ N times and forms a bias matrix with dimension $R^{N\times C}$. $C$ is the dimension of hidden states.

With this formulation, we can write down similar update rules for RGCN-GRU, RGAT-LSTM, RGAT-GRU, as well as for any other graph neural network recurrent unit combinations. 

Note that for a large-scale graph with millions of nodes, training for the whole graph becomes unfeasible because of the memory limitation. We use the sampling method proposed in GraphSAGE \cite{hamilton2017inductive} for batched training. At each training iteration, we first sample a small batch of nodes $B_0$ and then recursively expand $B_l$ to $B_{l+1}$ by sampling $S_l$ neighbourhood nodes of $B_l$. With a GNN of $M$ layers, we get a hierarchy of nodes: $B_0,B_1,...,B_M$. Representations of target nodes $B_0$ are updated by aggregating node states from the bottom layer $B_M$ to the upper layer $B_0$.

\subsection{Supervised Learning}
Given a final representation $h_i$ for node $v_i$, we first project $h_i$ into the classification space and get an output $o_i$.
\begin{align}
    o_i = W_o h_i +b_o
\end{align}
where $W_o\in R^{ |Y|\times C}, b_o\in R^{|Y|}$ are a projection matrix and a bias vector. $|Y|$ is the number of target classes. 

In a multi-label classification case, where labels are not mutually exclusive, the loss function for node $v_i$ is written as 
\begin{align}
    loss = \frac{1}{|Y|}\sum_{j=1}^{|Y|} [y_{ij}log(\sigma(o_{ij}))+(1-y_{ij})log(1-\sigma(o_{ij}))]
\end{align}
where $y_{ij}\in \{0,1\}$ is the label for class $j$.

In a multi-class classification setting, where labels are mutually exclusive, the loss is the cross-entropy loss after softmax normalization.
\begin{align}
    loss = -y_{ij}log(\frac{exp(o_{ij})}{\sum_{j}exp(o_{ij})})
\end{align}

\subsection{Unsupervised Learning}

Following previous work \cite{tang2015line,hamilton2017inductive}, in the unsupervised setting, we learn node representations by network modeling. Specifically, given a node $v_i$ with representation $h_i$, the goal is to optimize the probability of observing a context node $v_j$:
\begin{align}
    p(v_j|v_i) = \frac{exp(h_j^Th_i)}{\sum_{k=1}^N exp(h_k^T h_i)}
    \label{cond_prob}
\end{align}
where context node $v_j$ is generated by a random walk starting from node $v_i$.

Optimizing the conditional probability in Eq. \ref{cond_prob} for all context node pairs implies that nodes in proximity should have similar hidden states. In practise, optimizing Eq. \ref{cond_prob} is computationally expensive, since there are $N$ nodes involved in the denominator. So we use negative sampling \cite{mikolov2013distributed} to approximate it and the objective becomes:
\begin{align}
    log \sigma(h_j^Th_i)+\sum_{k=1}^K E_{v_k\sim P_N(v)}[log\sigma(-h_k^Th_i)]  
\end{align}
The task turns into distinguishing the context node $v_j$ from $K$ randomly sampled negative nodes. We use uniform distribution $P_N(v)$ here. To further reduce memory consumption in our batched training, nodes in one batch share the same set of negative nodes, which works well in practise.  

\begin{table*}[!th]
\caption{Comparative evaluation results for three datasets. We report micro-averaged F1 scores. ``-'' signifies no results are published for the given setting. }
\label{rgnn_compare}
\centering
\begin{tabular}{cclcclcc}
\hline\hline
\multirow{2}{*}{Methods} & Pubmed         &  & \multicolumn{2}{c}{Reddit}      &  & \multicolumn{2}{c}{PPI}         \\ \cline{2-2} \cline{4-5} \cline{7-8} 
                         & Sup. F1        &  & Unsup. F1      & Sup. F1        &  & Unsup. F1      & Sup. F1        \\ \hline
GCN                      & 0.875          &  & -              & 0.930          &  & -              & 0.865          \\
FastGCN                  & 0.880          &  & -              & 0.937          &  & -              & 0.607             \\
GAT                      & 0.883          &  & -              & 0.950          &  & -              & 0.973          \\
GraphSAGE-GCN            & 0.849          &  & 0.908          & 0.930          &  & 0.465          & 0.500          \\
GraphSAGE-mean           & 0.888          &  & 0.897          & 0.950          &  & 0.486          & 0.598          \\ \hline
RGCN-LSTM                & \textbf{0.908} &  & 0.919          & 0.963          &  & 0.791          & 0.992          \\
RGCN-GRU                 & 0.900          &  & 0.915          & \textbf{0.964} &  & 0.765         & 0.991          \\
RGAT-LSTM                & 0.905          &  & \textbf{0.921} & \textbf{0.964} &  & \textbf{0.806} & \textbf{0.994} \\
RGAT-GRU                 & 0.902          &  & 0.913          & \textbf{0.964} &  & 0.791         & 0.994          \\ \hline\hline
\end{tabular}
\end{table*}

\section{Experiments}

\subsection{Datasets}

We adopt three commonly used benchmark datasets in our experiments. A summary of these datasets is shown in Table \ref{graph_data}.

\begin{table}[!h]
\caption{Statistics of three datasets}
\label{graph_data}
\centering
\begin{tabular}{ccccc}
\hline\hline
Data   & \# Nodes & \# Edges      & \# Features & \# Classes \\ \hline
Pubmed & 19, 717 (1 graph) & 44, 338      & 500        & 3         \\
Reddit & 232,965 (1 graph) & 11, 606, 919 & 602        & 41        \\
PPI    & 56,944 (24 graph) & 818,716      & 50         & 121       \\ \hline\hline
\end{tabular}
\end{table}

\textbf{Pubmed} is a citation dataset introduced by \cite{sen2008collective}. Nodes represent academic papers within the Pubmed database and links are citations between papers. Node features are sparse bag-of-words representations of papers. Labels are the categories of these papers. Following \cite{chen2018fastgcn}, we use all labeled training examples for training per the supervised learning scenario. Because of the sparsity of this graph, we only test models with the supervised setting on Pubmed.

\textbf{Reddit} is a social network dataset compiled in \cite{hamilton2017inductive}. It contains 232K Reddit posts as nodes. If the same user comments on two posts then there is a link between these two posts. Node features are generated from Glove word embeddings \cite{pennington2014glove}. The node label in this case is the ``subreddit'' a post belongs to. Because of the graph size, we apply batched training on this dataset.

\textbf{PPI} contains 24 protein-protein interaction graphs, with each graph corresponding to a different human tissue \cite{zitnik2017predicting}. Node features include positional gene sets, motif gene sets, and immunological signatures. Node labels are protein roles in terms of their cellular functions. Following \cite{hamilton2017inductive}, we train all models on 20 graphs, validate and test on 2 graphs each. It validates the generalizing performance across graphs.

\subsection{Experimental Setup}
\textbf{Supervised learning} We set dimensions of hidden states as 64, 600, and 1024 for Pubmed, Reddit, and PPI respectively. For GAT based models, we use 8 heads for Pubmed, 5 heads for Reddit, 4 heads for PPI. We apply dropout \cite{srivastava2014dropout} on the input features for Pubmed, PPI with dropout rate 0.2. We apply two-layer GNN models for Pubmed and Reddit, and three-layer ones for PPI. Each layer in GNNs is followed by an exponential linear unit (ELU) nonlinearity \cite{clevert2015fast}. Models are trained with Adam optimizer \cite{kingma2014adam} with an initial learning rate of 0.01 for Pubmed, and 0.001 for other datasets. We apply batched training on the Reddit dataset with neighborhood sample sizes $S_1=25$ and $S_2 = 10$. The batch size is 128. Because of the dataset size, we run models 5 times and report an average performance for Pubmed and PPI.

\textbf{Unsupervised learning}: In the unsupervised setting, we use two-layer GNN models. The negative sampling size $K$ is 10. Random walk lengths for PPI and Reddit are 2 and 3 respectively. Other hyperparameter settings are the same with supervised learning, except for that we have not applied dropout here. After we get the node representations with unsupervised learning, we use representations of training nodes to train a downstream linear classifier same as GraphSAGE \cite{hamilton2017inductive}.

In both learning scenarios, we strictly follow the inductive learning setting where validation and test nodes are hidden during training. We ran our experiments on a linux machine with 4 NVIDIA Titan XP GPUs (12GB of RAM), one Intel Core i7-6850K CPU, 128GB of RAM. 

\subsection{Baseline Comparisons}

Results of comparative evaluation experiments are shown in Table \ref{rgnn_compare}. We evaluate RGCN and RGAT with LSTM/GRU units. In the supervised setting, we compare RGNN based models with various baselines --- GCN, FastFCN \cite{chen2018fastgcn}, GAT, and GraphSAGE models. In the unsupervised setting, we use GraphSAGEs as baselines.

\begin{figure*}[!t]
\centering
\begin{subfigure}{.5\textwidth}
  \centering
  \includegraphics[width=\linewidth]{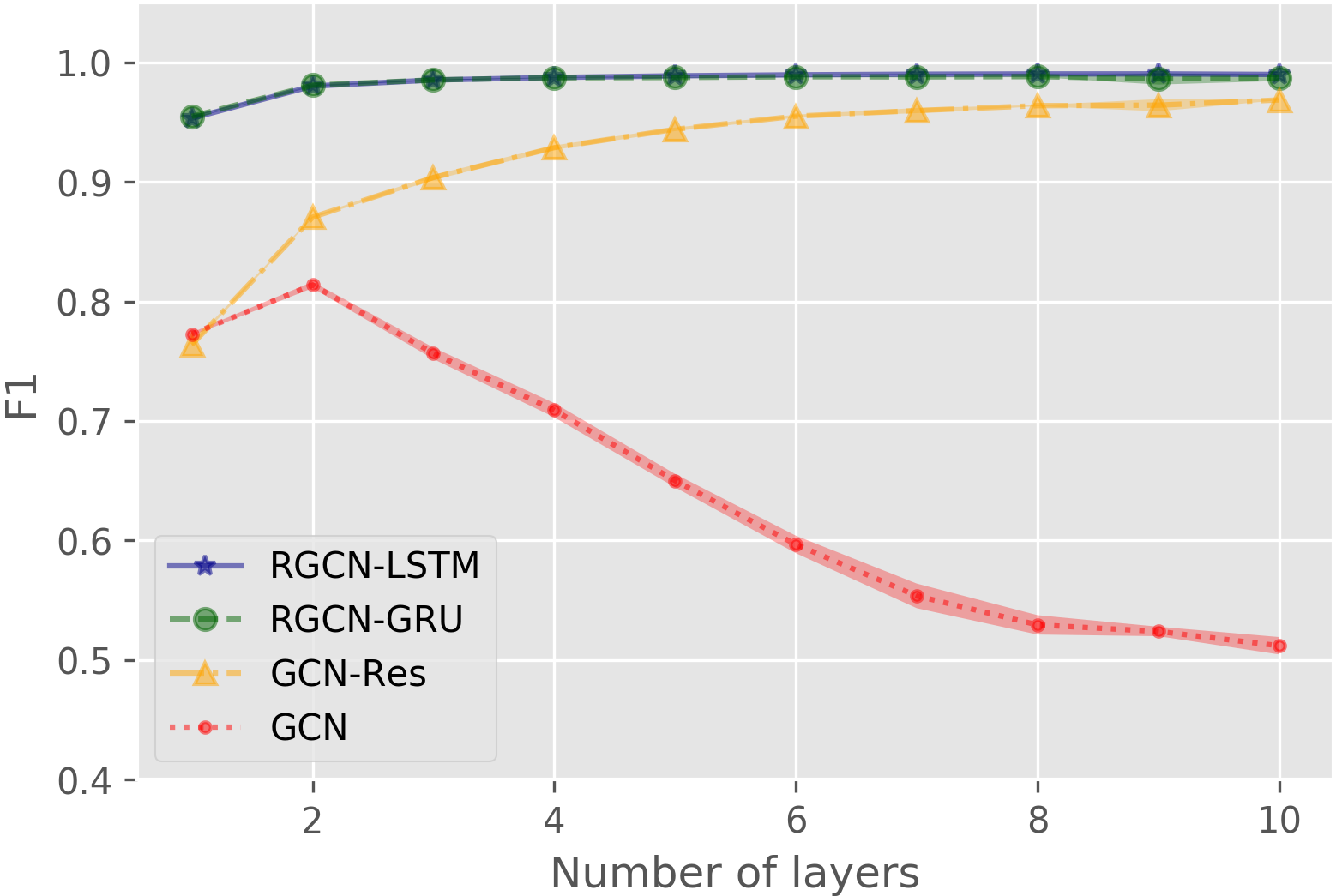}
  \caption{GCN based models on PPI}
   \label{depth_gcn_ppi}
\end{subfigure}%
\begin{subfigure}{.5\textwidth}
  \centering
  \includegraphics[width=\linewidth]{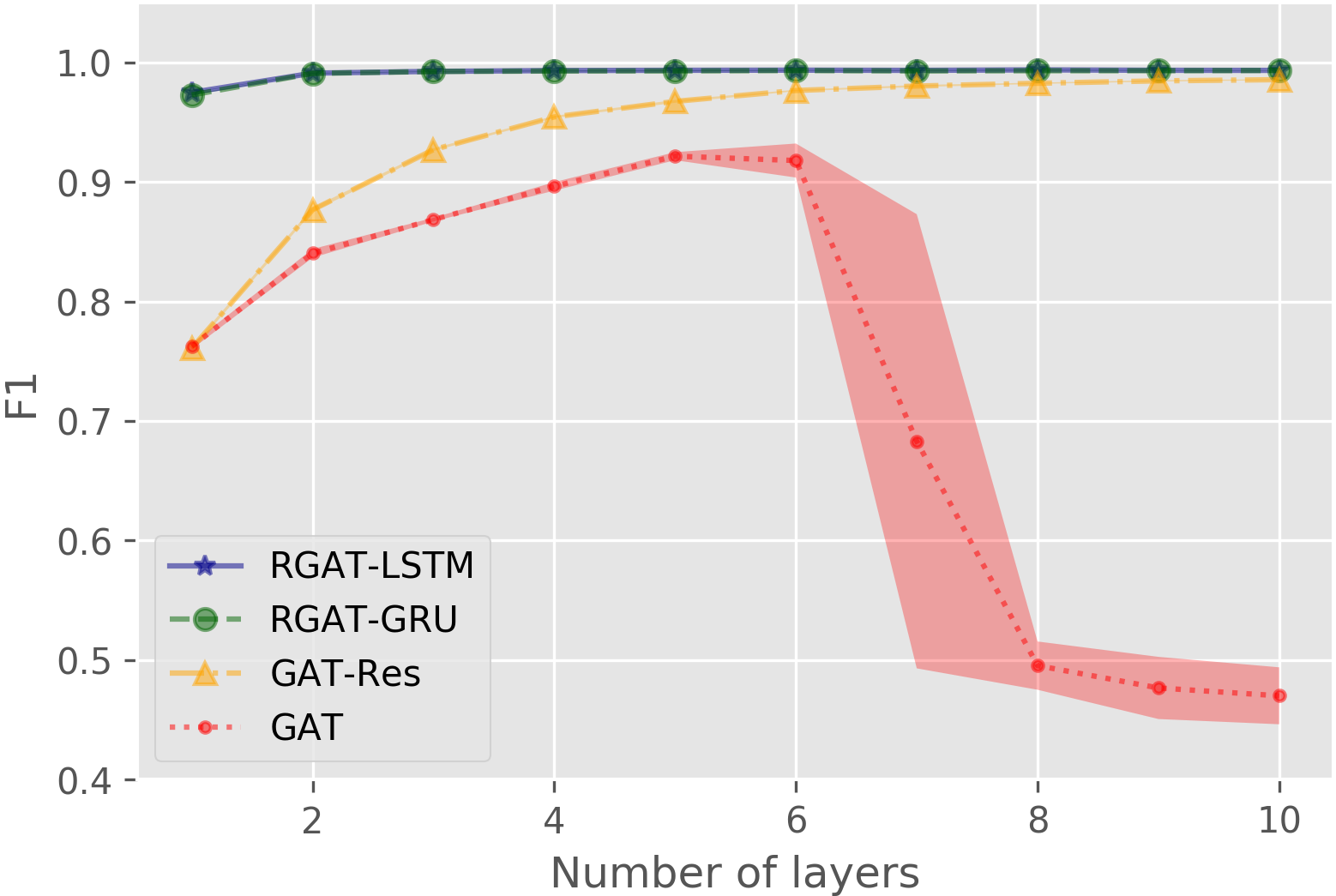}
  \caption{GAT based models on PPI}
  \label{depth_gat_ppi}
\end{subfigure}

\begin{subfigure}{.5\textwidth}
  \centering
  \includegraphics[width=\linewidth]{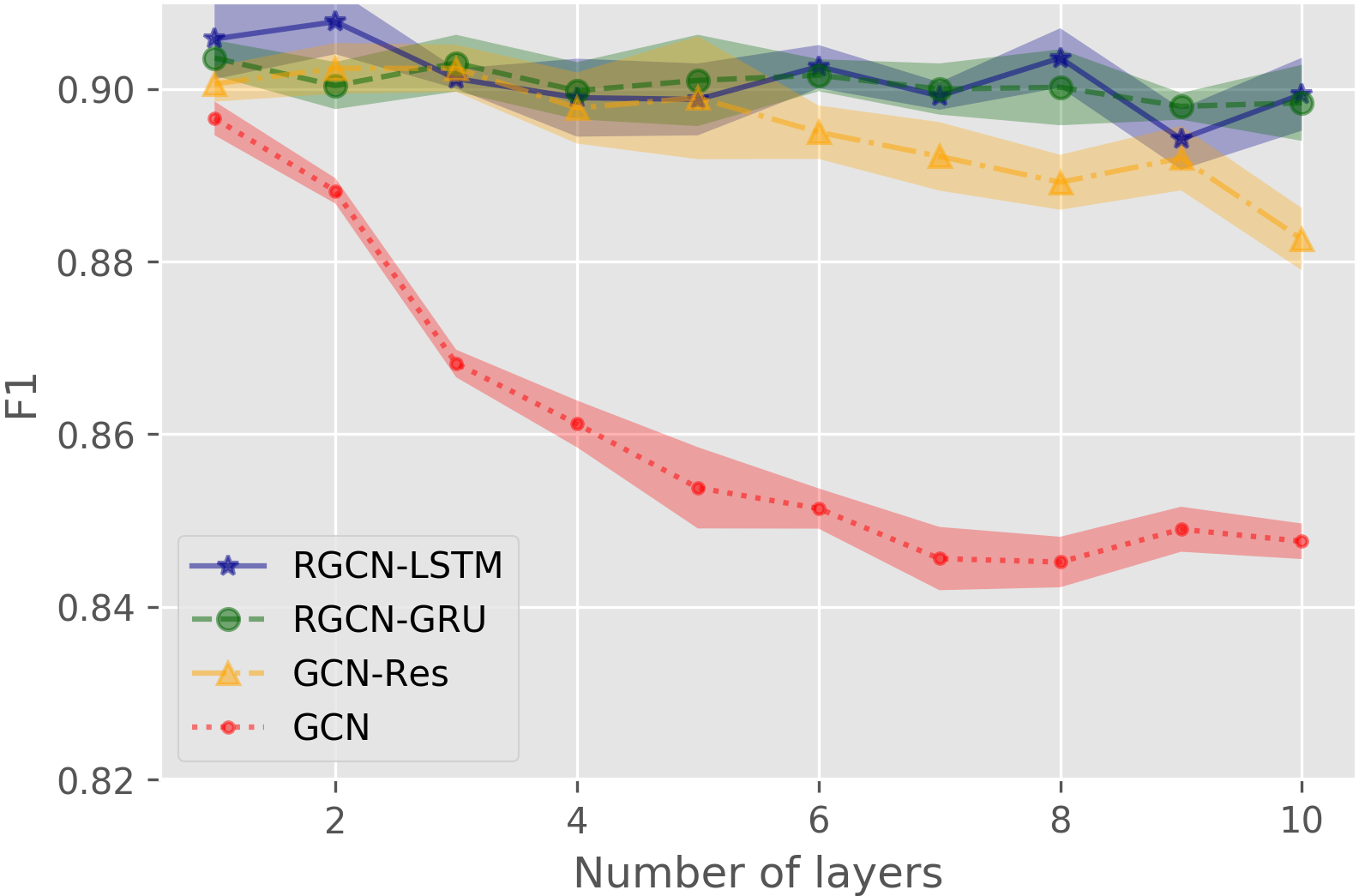}
  \caption{GCN based models on Pubmed}
  \label{depth_gcn_pubmed}
\end{subfigure}%
\begin{subfigure}{.5\textwidth}
  \centering
  \includegraphics[width=\linewidth]{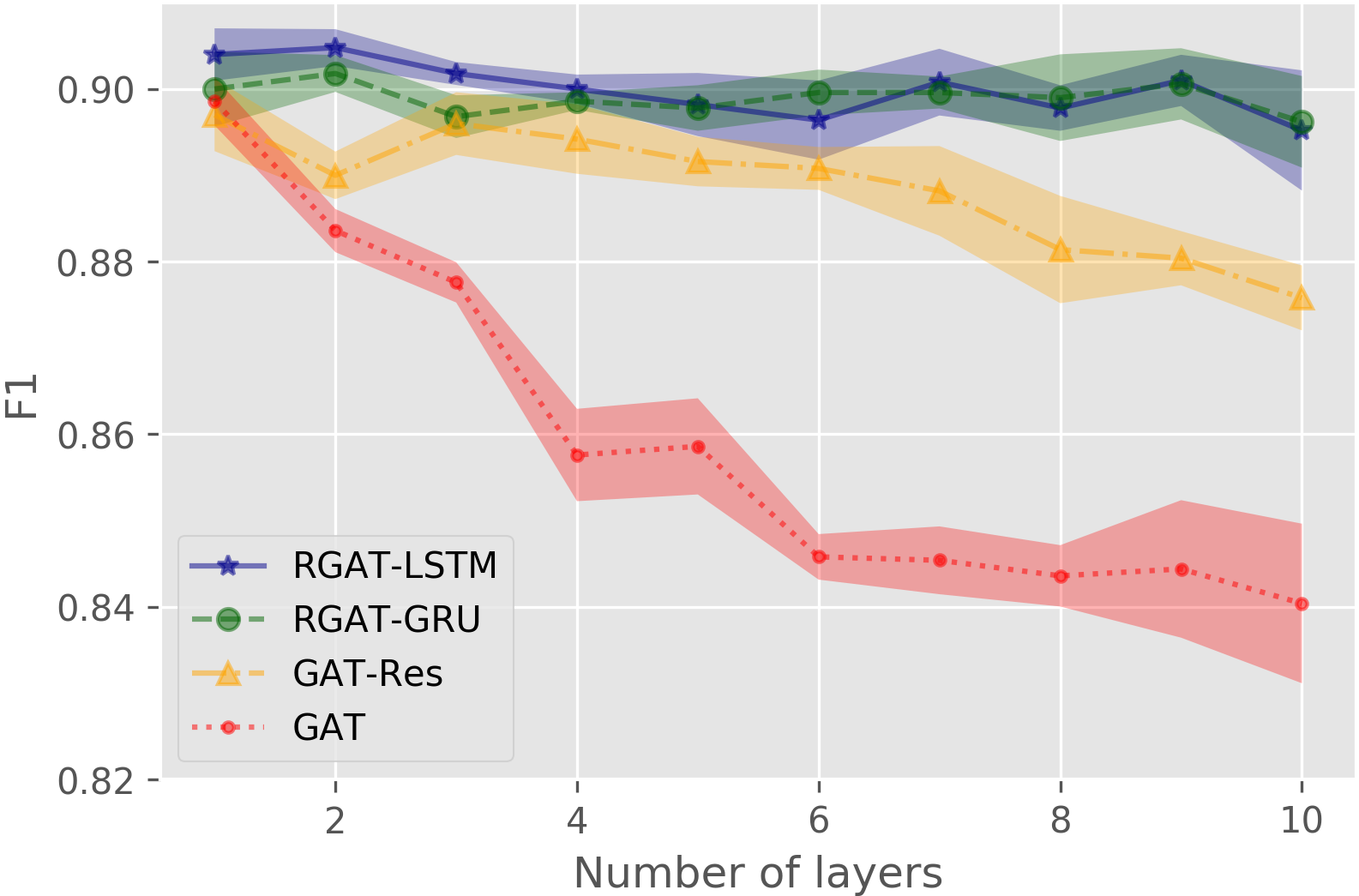}
  \caption{GAT based models on Pubmed}
  \label{depth_gat_pubmed}
\end{subfigure}
\caption{Influence of model depth (number of layers) on performance. Markers denote averaged micro-F1 scores on test dataset in 5 runs. Shaded areas represent standard deviations. We show results for RGNN with LSTM and GRU units, GNN with residual connections, and standard GNN models.}
\label{compare_depth}
\end{figure*}

Our results demonstrate that unifying recurrent units in modern GNN models can effectively improve state-of-the-art performance across all three datasets. In the supervised learning setting, we are able to improve GCNs by an absolute increase of 12.7\% on PPI. For Reddit and Pubmed, 2\% - 4\% improvement is achieved. Note that even Veli{\v{c}}kovi{\'c} et al. use residual connections for GAT on PPI dataset, their result is still lower than RGAT with recurrent units.

In the unsupervised setting, we observe similar improvement on Reddit and PPI datasets. Noticeably, RGNN based models perform much better on PPI than baselines under unsupervised learning. Our best model RGAT-LSTM achieves over 30\% improvement over GraphSAGE, which is even better than some baseline models with supervised signals. Comparing RGCN-LSTM with GraphSAGE-GCN, we can find the LSTM unit provides a significant gain on this task.

\subsection{Model Depth Analysis}

In this section, we investigate the influence of model depth (number of layers) on performance and compare the effects of adding recurrent units and residual connection. In this experiment, we use the same hyperparameter setting across all the base models. We run each method 5 times on PPI and Pubmed under supervised setting and report the average micro-F1 and the standard deviation in Figure \ref{compare_depth}. Because of the GPU memory limitation, we change the dimension of hidden states from 1024 to 512 on PPI in these experiments.  

As shown in Figure \ref{depth_gcn_ppi} and \ref{depth_gat_ppi}, on PPI dataset, GNNs with recurrent units can be easily extended to deeper models and perform much better than GNNs with residual connections (GNN-Res). A vanilla GCN degenerates quickly when the depth increases to 3 or higher. The GAT is better than the GCN, but it still fails when its depth goes beyond 6. Using residual connections does help GAT and GCN models to generalize to deeper models. However, performances of GNN models with residual connections are still worse than GNNs with recurrent units. RGCN models and RGAT models can quickly reach and maintain their optimal performances on PPI dataset. Although the difference between GAT-Res and RGAT models becomes less when depth gets larger, a 10-layer GAT-Res is still worse than 10-layer RGAT-LSTM/RGAT-GRU (0.985 versus 0.993/0.993). 

In Figure \ref{depth_gcn_pubmed} and \ref{depth_gat_pubmed}, for Pubmed dataset, best results are obtained with shallow models. Without residual connections or recurrent units, the performance of GNN models decreases with larger depth. GNN-Res models degenerate when the model is deeper than 5 layers. On the contrary, deep RGCN and RGAT still work similarly to shallow models, as LSTM and GRU successfully capture the long-term dependency.

\subsection{Perturbation Analysis}

\begin{figure*}[!t]
\centering
\begin{subfigure}{.5\textwidth}
  \centering
  \includegraphics[width=\linewidth]{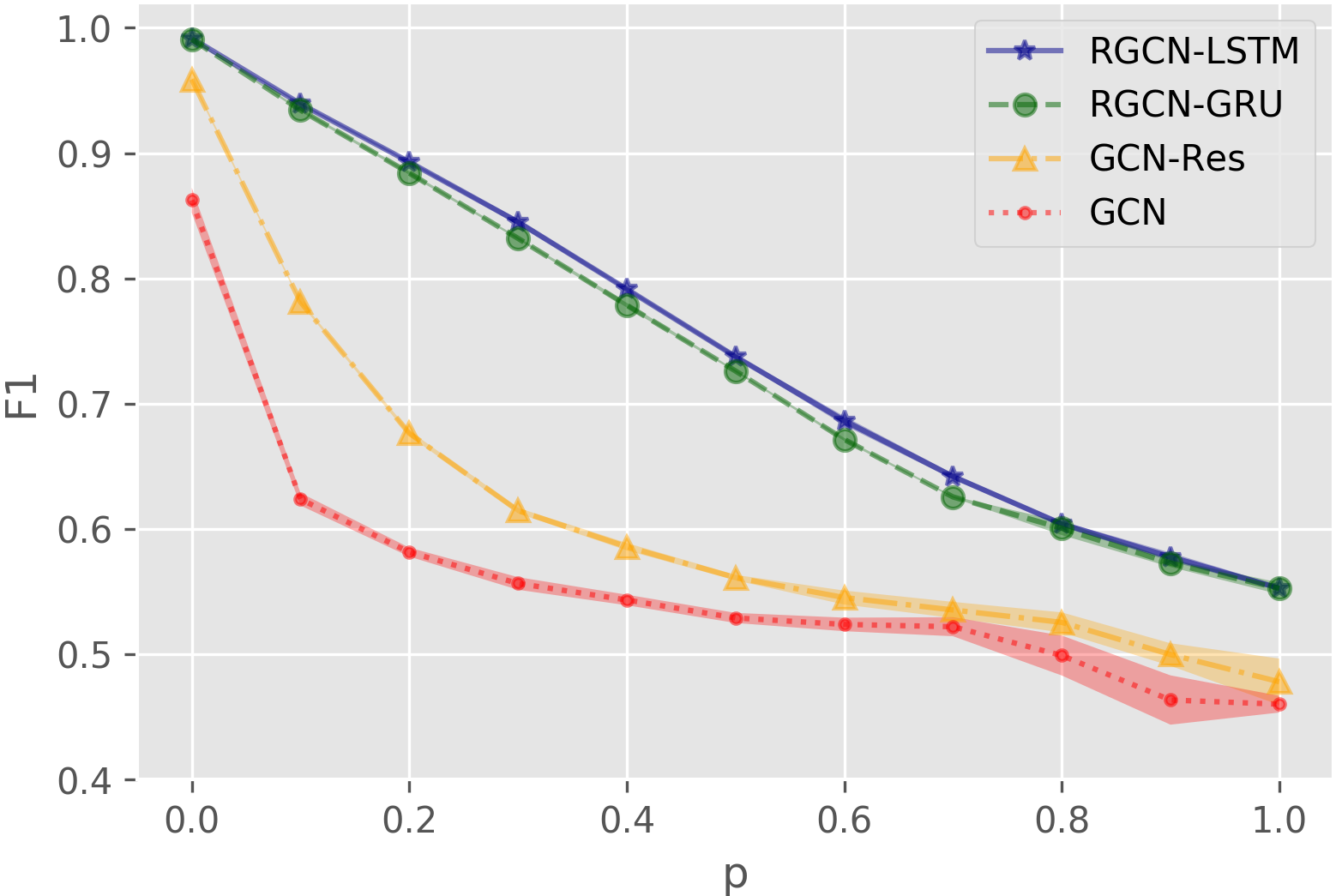}
  \caption{GCN based models with noisy graphs}
  \label{noisy_graph_gcn}
\end{subfigure}%
\begin{subfigure}{.5\textwidth}
  \centering
  \includegraphics[width=\linewidth]{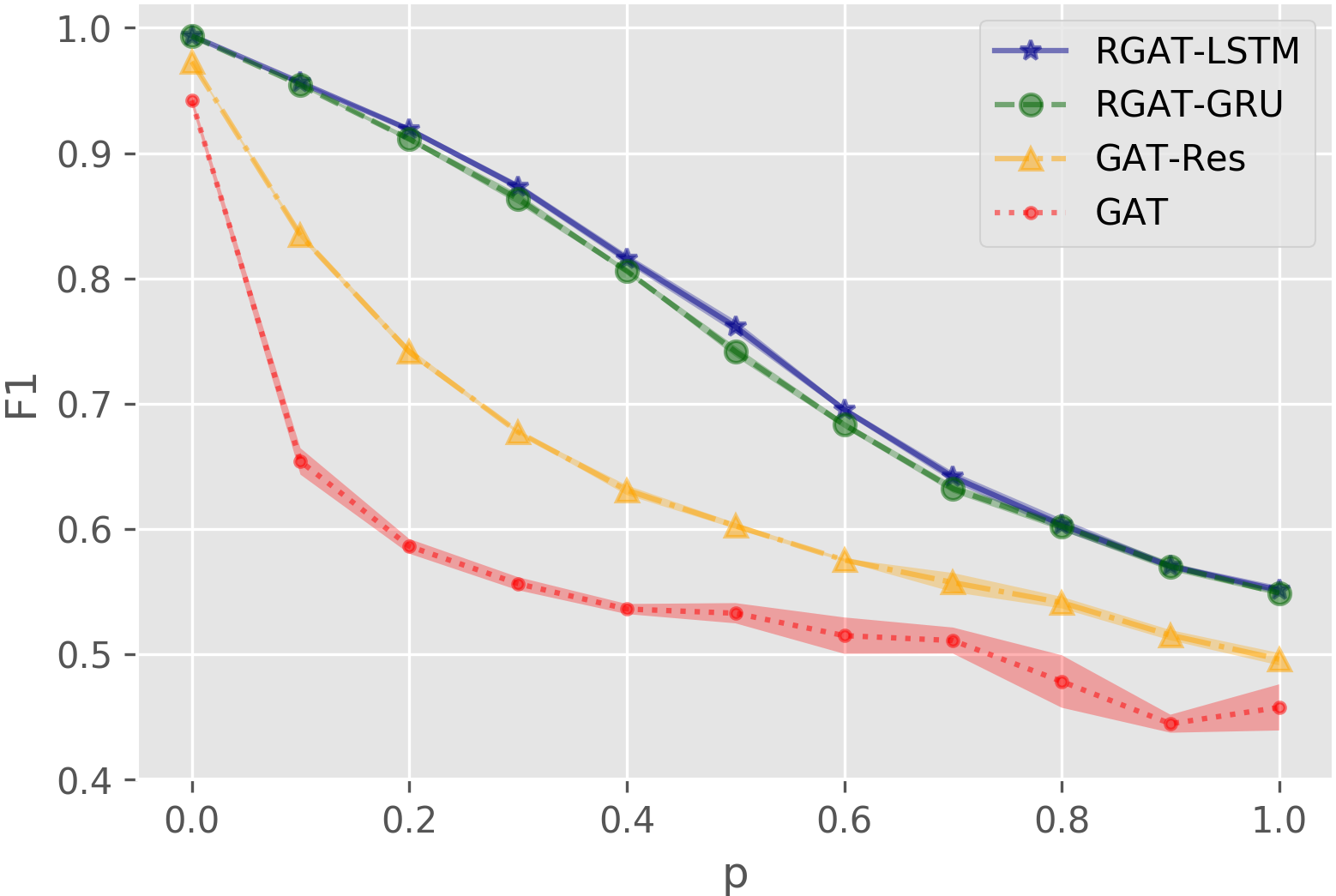}
  \caption{GAT based models with noisy graphs}
  \label{noisy_graph_gat}
\end{subfigure}
\begin{subfigure}{.5\textwidth}
  \centering
  \includegraphics[width=\linewidth]{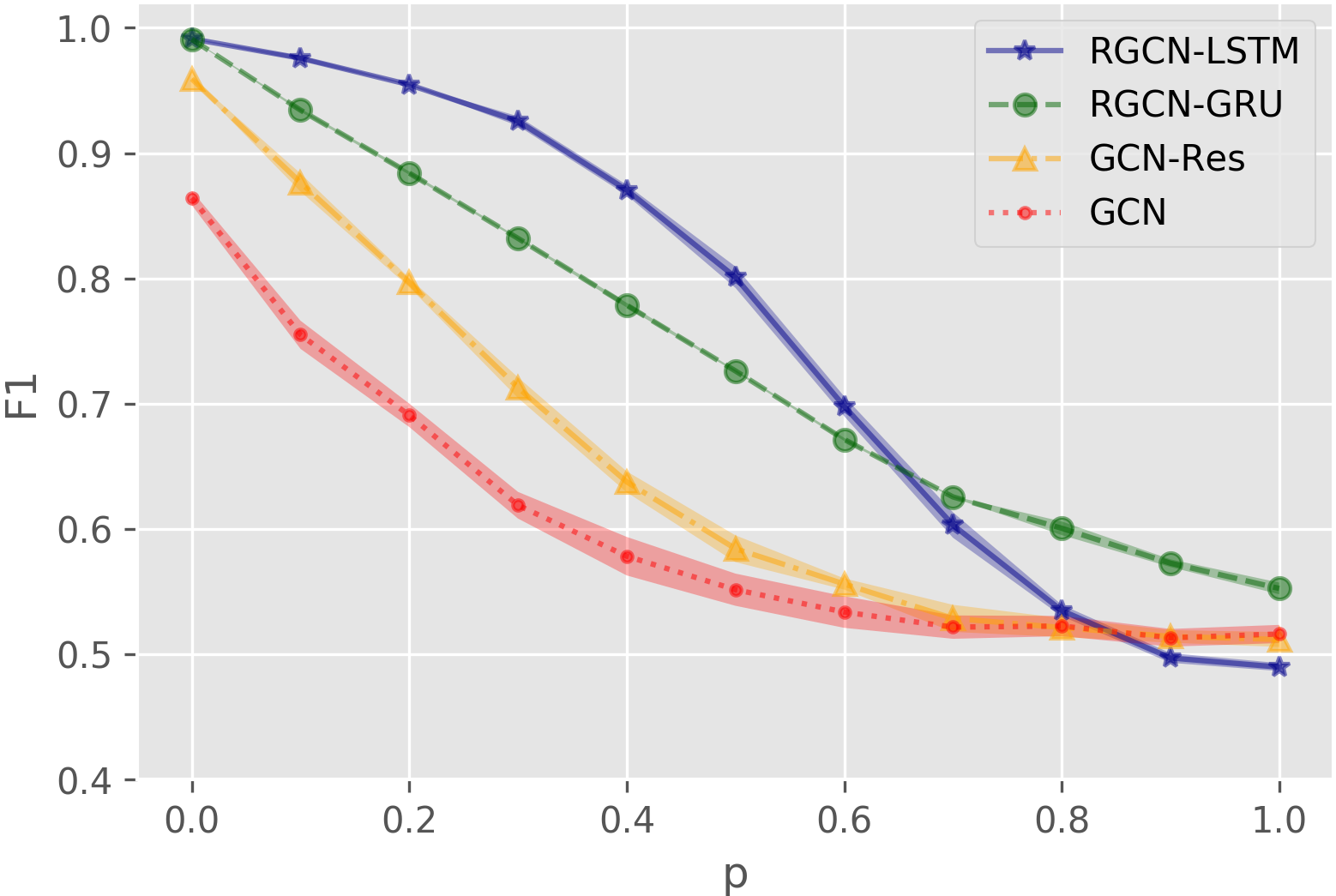}
  \caption{GCN based models with noisy features}
  \label{noisy_feature_gcn}
\end{subfigure}%
\begin{subfigure}{.5\textwidth}
  \centering
  \includegraphics[width=\linewidth]{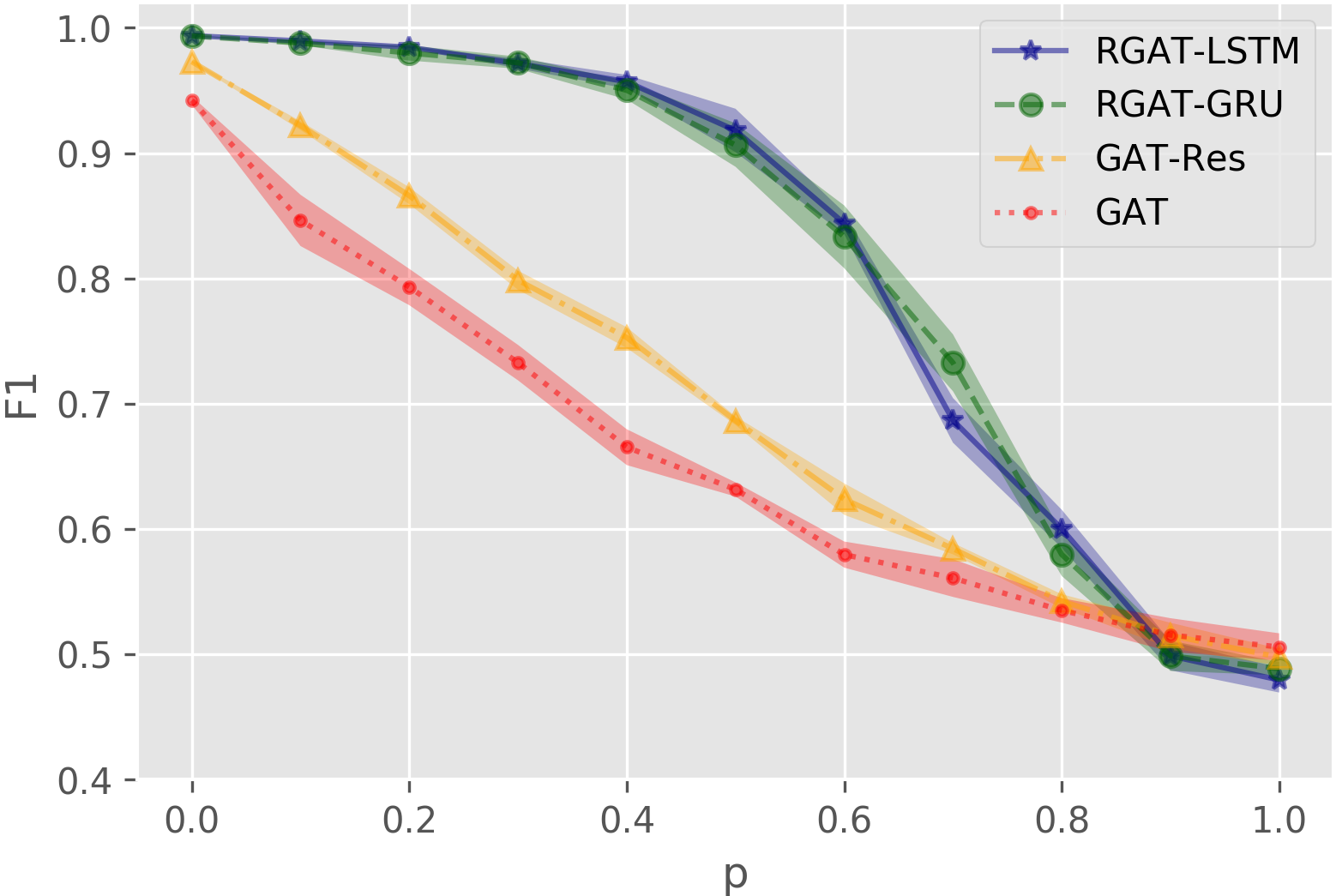}
  \caption{GAT based models with noisy features}
  \label{noisy_feature_gat}
\end{subfigure}
\caption{Perturbation analysis. Markers denote averaged micro-F1 scores on test graphs of PPI in 5 runs. Shaded areas represent standard deviations. We show results for RGNN with LSTM and GRU units, GNN with residual connections, and standard GNN models. }
\label{noisy_graph}
\end{figure*}

For many real-world problems, we do not have access to accurate information about the graph. Many times, there is noisy information in the graph structure and nodes' local features. In this section, we perform a perturbation study where we compare 3-layer RGNN models against 3-layer GNNs with imperfect information on PPI dataset under supervised learning. 

In the first noisy graph scenario, we cut an edge with probability $p$ and connect two randomly selected nodes. When $p$ equals to $1$, the reconstructed graph turns to be a random graph with the same graph density. We measure the performance of RGNNs, GNNs with residual connections, and GNNs under various probability $p$. In Figure \ref{noisy_graph_gcn} and \ref{noisy_graph_gat}, we observe that RGNN models with LSTM and GRU units are more robust to noisy graph information, which shows that gates in these two RNN units are helpful for capturing important information and avoiding noisy graph information. RGNNs with LSTM units generally work better than RGNN with GRU units in this case.

In the second noisy feature scenario, for each node, we randomly mutate its local features with probability $p$, where we replace its features with Gaussian noises draw from $N(0,1)$. As shown in Figure \ref{noisy_feature_gcn} and \ref{noisy_feature_gat}, RGNN models have a better capability of distinguishing noisy features than GNNs with residual connections. The performance of RGCN-LSTM is generally better than RGCN-GRU, but it decreases faster than RGCN-GRU in extreme cases ($p>0.7$). RGAT-LSTM and RGAT-GRU work similarly and they both outperform GAT-Res and GAT in a large margin.

\section{Related Work}
Many network embedding methods have been proposed to map nodes into low dimensional embedding vectors. Nodes with close proximity would have similar embedding vectors. Typical examples include DeepWalk \cite{perozzi2014deepwalk}, node2vec \cite{grover2016node2vec}, LINE \cite{tang2015line}. DeepWalk utilizes random walk to sample context nodes and use SkipGram \cite{mikolov2013distributed} to model the occurrences among node pairs. node2vec uses a similar strategy, but with a different biased random walk method. As summarized in \cite{qiu2018network}, these mentioned embedding methods can be unified into a matrix factorization framework. Because these methods have to train embedding vectors for all the nodes, they cannot be easily generalized to unseen nodes during training.

In recent literature, graph neural networks show great success in learning node representations \cite{hamilton2017inductive}. Gated Graph Neural Networks (GGNN) \cite{li2015gated} extend vanilla graph neural networks by adding gating mechanism and relaxes the contraction map assumption. GGNN can be viewed a special case of RGNN. However, a key difference here is that a GGNN shares parameter across layers, which may limits the expressive power especially for large graphs with rich features. Besides, our framework is more flexible to accommodate all types of graph neural network and recurrent unit combinations like LSTM, IndRNN \cite{li2018independently}, and SRU \cite{lei2017training}. There is also some other work incorporating GNNs with GRU units \cite{li2017diffusion,zhang2018gaan}. Li et al. use GRU to model the temporal dependency in a sequence of traffic network. The matrix multiplication in GRU is replaced by the diffusion convolution operation on the traffic network at each time step \cite{li2017diffusion}. Similarly, Seo et al. replace the matrix multiplication in LSTM with graph convolution to model a temporal sequence of graphs \cite{seo2018structured}.
However, all these models are designed to handle sequence input and the recurrent unit is used to model sequence dependency through time, while the presented RGNN model class uses recurrent units to capture important information across GNN layers on static graphs. One similar work also tries to embed LSTM into the propagation process of a graph convolution neural network. However, it is still unknown that why does such modification work, how does it compare to other method like residual connection, and can we generalize this method to other GNN classes.

\section{Conclusion}

In this paper, we present a general graph neural network class named recurrent graph neural network for inductive graph representation learning. It combines graph neural networks with recurrent units to avoid noisy neighbourhood information across layers. Compared to previous methods, the presented RGNN models establish new state-of-the-art results on three benchmark datasets under both the supervised setting and the unsupervised setting. In our experiments, we systematically evaluate the effect of adding recurrent units. Our results demonstrate that GNN models with recurrent units are much easier to extend to deeper models than GNN models with residual connections. In our further analyses, we show RGNN models are more robust to noisy information from graph structure as well as local features.

\bibliographystyle{IEEEtran}
\bibliography{./reference}

\begin{thebibliography}{10}
\providecommand{\url}[1]{#1}
\csname url@samestyle\endcsname
\providecommand{\newblock}{\relax}
\providecommand{\bibinfo}[2]{#2}
\providecommand{\BIBentrySTDinterwordspacing}{\spaceskip=0pt\relax}
\providecommand{\BIBentryALTinterwordstretchfactor}{4}
\providecommand{\BIBentryALTinterwordspacing}{\spaceskip=\fontdimen2\font plus
\BIBentryALTinterwordstretchfactor\fontdimen3\font minus
  \fontdimen4\font\relax}
\providecommand{\BIBforeignlanguage}[2]{{%
\expandafter\ifx\csname l@#1\endcsname\relax
\typeout{** WARNING: IEEEtran.bst: No hyphenation pattern has been}%
\typeout{** loaded for the language `#1'. Using the pattern for}%
\typeout{** the default language instead.}%
\else
\language=\csname l@#1\endcsname
\fi
#2}}
\providecommand{\BIBdecl}{\relax}
\BIBdecl

\bibitem{perozzi2014deepwalk}
B.~Perozzi, R.~Al-Rfou, and S.~Skiena, ``Deepwalk: Online learning of social
  representations,'' in \emph{Proceedings of the 20th ACM SIGKDD international
  conference on Knowledge discovery and data mining}.\hskip 1em plus 0.5em
  minus 0.4em\relax ACM, 2014, pp. 701--710.

\bibitem{grover2016node2vec}
A.~Grover and J.~Leskovec, ``node2vec: Scalable feature learning for
  networks,'' in \emph{Proceedings of the 22nd ACM SIGKDD international
  conference on Knowledge discovery and data mining}.\hskip 1em plus 0.5em
  minus 0.4em\relax ACM, 2016, pp. 855--864.

\bibitem{tang2015line}
J.~Tang, M.~Qu, M.~Wang, M.~Zhang, J.~Yan, and Q.~Mei, ``Line: Large-scale
  information network embedding,'' in \emph{Proceedings of the 24th
  international conference on world wide web}.\hskip 1em plus 0.5em minus
  0.4em\relax International World Wide Web Conferences Steering Committee,
  2015, pp. 1067--1077.

\bibitem{bhagat2011node}
S.~Bhagat, G.~Cormode, and S.~Muthukrishnan, ``Node classification in social
  networks,'' in \emph{Social network data analytics}.\hskip 1em plus 0.5em
  minus 0.4em\relax Springer, 2011, pp. 115--148.

\bibitem{lu2011link}
L.~L{\"u} and T.~Zhou, ``Link prediction in complex networks: A survey,''
  \emph{Physica A: statistical mechanics and its applications}, vol. 390,
  no.~6, pp. 1150--1170, 2011.

\bibitem{kipf2016semi}
T.~N. Kipf and M.~Welling, ``Semi-supervised classification with graph
  convolutional networks,'' \emph{arXiv preprint arXiv:1609.02907}, 2016.

\bibitem{hamilton2017inductive}
W.~Hamilton, Z.~Ying, and J.~Leskovec, ``Inductive representation learning on
  large graphs,'' in \emph{Advances in Neural Information Processing Systems},
  2017, pp. 1024--1034.

\bibitem{he2016deep}
K.~He, X.~Zhang, S.~Ren, and J.~Sun, ``Deep residual learning for image
  recognition,'' in \emph{Proceedings of the IEEE conference on computer vision
  and pattern recognition}, 2016, pp. 770--778.

\bibitem{velivckovic2017graph}
P.~Veli{\v{c}}kovi{\'c}, G.~Cucurull, A.~Casanova, A.~Romero, P.~Lio, and
  Y.~Bengio, ``Graph attention networks,'' \emph{arXiv preprint
  arXiv:1710.10903}, 2017.

\bibitem{hochreiter1997long}
S.~Hochreiter and J.~Schmidhuber, ``Long short-term memory,'' \emph{Neural
  computation}, vol.~9, no.~8, pp. 1735--1780, 1997.

\bibitem{cho2014learning}
K.~Cho, B.~Van~Merri{\"e}nboer, C.~Gulcehre, D.~Bahdanau, F.~Bougares,
  H.~Schwenk, and Y.~Bengio, ``Learning phrase representations using rnn
  encoder-decoder for statistical machine translation,'' \emph{arXiv preprint
  arXiv:1406.1078}, 2014.

\bibitem{sundermeyer2012lstm}
M.~Sundermeyer, R.~Schl{\"u}ter, and H.~Ney, ``Lstm neural networks for
  language modeling,'' in \emph{Thirteenth annual conference of the
  international speech communication association}, 2012.

\bibitem{gers2002learning}
F.~A. Gers, N.~N. Schraudolph, and J.~Schmidhuber, ``Learning precise timing
  with lstm recurrent networks,'' \emph{Journal of machine learning research},
  vol.~3, no. Aug, pp. 115--143, 2002.

\bibitem{chung2014empirical}
J.~Chung, C.~Gulcehre, K.~Cho, and Y.~Bengio, ``Empirical evaluation of gated
  recurrent neural networks on sequence modeling,'' \emph{arXiv preprint
  arXiv:1412.3555}, 2014.

\bibitem{scarselli2009graph}
F.~Scarselli, M.~Gori, A.~C. Tsoi, M.~Hagenbuchner, and G.~Monfardini, ``The
  graph neural network model,'' \emph{IEEE Transactions on Neural Networks},
  vol.~20, no.~1, pp. 61--80, 2009.

\bibitem{vaswani2017attention}
A.~Vaswani, N.~Shazeer, N.~Parmar, J.~Uszkoreit, L.~Jones, A.~N. Gomez,
  {\L}.~Kaiser, and I.~Polosukhin, ``Attention is all you need,'' in
  \emph{Advances in Neural Information Processing Systems}, 2017, pp.
  5998--6008.

\bibitem{mikolov2013distributed}
T.~Mikolov, I.~Sutskever, K.~Chen, G.~S. Corrado, and J.~Dean, ``Distributed
  representations of words and phrases and their compositionality,'' in
  \emph{Advances in neural information processing systems}, 2013, pp.
  3111--3119.

\bibitem{sen2008collective}
P.~Sen, G.~Namata, M.~Bilgic, L.~Getoor, B.~Galligher, and T.~Eliassi-Rad,
  ``Collective classification in network data,'' \emph{AI magazine}, vol.~29,
  no.~3, pp. 93--93, 2008.

\bibitem{chen2018fastgcn}
J.~Chen, T.~Ma, and C.~Xiao, ``Fastgcn: fast learning with graph convolutional
  networks via importance sampling,'' \emph{arXiv preprint arXiv:1801.10247},
  2018.

\bibitem{pennington2014glove}
J.~Pennington, R.~Socher, and C.~Manning, ``Glove: Global vectors for word
  representation,'' in \emph{Proceedings of the 2014 conference on empirical
  methods in natural language processing (EMNLP)}, 2014, pp. 1532--1543.

\bibitem{zitnik2017predicting}
M.~Zitnik and J.~Leskovec, ``Predicting multicellular function through
  multi-layer tissue networks,'' \emph{Bioinformatics}, vol.~33, no.~14, pp.
  i190--i198, 2017.

\bibitem{srivastava2014dropout}
N.~Srivastava, G.~Hinton, A.~Krizhevsky, I.~Sutskever, and R.~Salakhutdinov,
  ``Dropout: a simple way to prevent neural networks from overfitting,''
  \emph{The Journal of Machine Learning Research}, vol.~15, no.~1, pp.
  1929--1958, 2014.

\bibitem{clevert2015fast}
D.-A. Clevert, T.~Unterthiner, and S.~Hochreiter, ``Fast and accurate deep
  network learning by exponential linear units (elus),'' \emph{arXiv preprint
  arXiv:1511.07289}, 2015.

\bibitem{kingma2014adam}
D.~P. Kingma and J.~Ba, ``Adam: A method for stochastic optimization,''
  \emph{arXiv preprint arXiv:1412.6980}, 2014.

\bibitem{qiu2018network}
J.~Qiu, Y.~Dong, H.~Ma, J.~Li, K.~Wang, and J.~Tang, ``Network embedding as
  matrix factorization: Unifying deepwalk, line, pte, and node2vec,'' in
  \emph{Proceedings of the Eleventh ACM International Conference on Web Search
  and Data Mining}.\hskip 1em plus 0.5em minus 0.4em\relax ACM, 2018, pp.
  459--467.

\bibitem{li2015gated}
Y.~Li, D.~Tarlow, M.~Brockschmidt, and R.~Zemel, ``Gated graph sequence neural
  networks,'' \emph{arXiv preprint arXiv:1511.05493}, 2015.

\bibitem{li2018independently}
S.~Li, W.~Li, C.~Cook, C.~Zhu, and Y.~Gao, ``Independently recurrent neural
  network (indrnn): Building a longer and deeper rnn,'' in \emph{Proceedings of
  the IEEE Conference on Computer Vision and Pattern Recognition}, 2018, pp.
  5457--5466.

\bibitem{lei2017training}
T.~Lei, Y.~Zhang, and Y.~Artzi, ``Training rnns as fast as cnns,'' \emph{arXiv
  preprint arXiv:1709.02755}, 2017.

\bibitem{li2017diffusion}
Y.~Li, R.~Yu, C.~Shahabi, and Y.~Liu, ``Diffusion convolutional recurrent
  neural network: Data-driven traffic forecasting,'' \emph{arXiv preprint
  arXiv:1707.01926}, 2017.

\bibitem{zhang2018gaan}
J.~Zhang, X.~Shi, J.~Xie, H.~Ma, I.~King, and D.-Y. Yeung, ``Gaan: Gated
  attention networks for learning on large and spatiotemporal graphs,''
  \emph{arXiv preprint arXiv:1803.07294}, 2018.

\bibitem{seo2018structured}
Y.~Seo, M.~Defferrard, P.~Vandergheynst, and X.~Bresson, ``Structured sequence
  modeling with graph convolutional recurrent networks,'' in
  \emph{International Conference on Neural Information Processing}.\hskip 1em
  plus 0.5em minus 0.4em\relax Springer, 2018, pp. 362--373.

\end{thebibliography}

\end{document}